\documentclass[8pt, a4paper, twocolumn]{extarticle}

\usepackage{amsmath}
\usepackage{geometry}
\usepackage[sort, numbers]{natbib}
\usepackage{siunitx}
\usepackage{todonotes}

\newcommand{\bs}[1]{\boldsymbol{#1}}

\newcommand{\Cm}{C_\mathrm{m}}

\newcommand{\El}{E_\mathrm{l}}
\newcommand{\Erev}{E^\mathrm{rev}}
\newcommand{\Ereve}{E^\mathrm{rev}_\mathrm{exc}}
\newcommand{\Erevi}{E^\mathrm{rev}_\mathrm{inh}}
\newcommand{\expect}[1]{\left\langle#1\right\rangle}
\newcommand{\gl}{g_\mathrm{l}}
\newcommand{\gtot}{g_\mathrm{tot}}

\newcommand{\Isyn}{I^\mathrm{syn}}
\newcommand{\non}{\mathrm{\setminus}}
\newcommand{\pB}{p_\mathrm{B}}

\newcommand{\PSP}{P\!S\!P}
\newcommand{\T}{\mathrm{T}}
\newcommand{\taueff}{\tau_\mathrm{eff}}
\newcommand{\taufac}{\tau_\mathrm{fac}}
\newcommand{\taum}{\tau_\mathrm{m}}
\newcommand{\taurec}{\tau_\mathrm{rec}}
\newcommand{\tauref}{\tau_\mathrm{ref}}
\newcommand{\tausyn}{\tau_\mathrm{syn}}
\newcommand{\tb}[1]{\textbf{#1}}

\newcommand{\uf}{u^\mathrm{f}}

\newcommand{\uthr}{\vartheta}

\title{\Huge Spiking neurons with short-term synaptic plasticity\\form superior generative networks}

\author{
    Luziwei Leng\textsuperscript{a,1}, Roman Martel\textsuperscript{a}, Oliver Breitwieser\textsuperscript{a}, Ilja Bytschok\textsuperscript{a},\\
    Walter Senn\textsuperscript{b}, Johannes Schemmel\textsuperscript{a}, Karlheinz Meier\textsuperscript{a}, Mihai A. Petrovici\textsuperscript{a,b,1,2}
    \vspace{10pt}\\
    \textsuperscript{a}\textit{Kirchhoff Institute for Physics, University of Heidelberg}\\
    \textsuperscript{b}\textit{Department of Physiology, University of Bern}\\
    \textsuperscript{1}\textit{Authors with equal contributions}\\
    \textsuperscript{1}\textit{Corresponding author}: mpedro@kip.uni-heidelberg.de 
    }

\date{\today}

\begin{document}

\maketitle

\section*{Significance statement}
Neural networks have long been used to solve various problems in machine learning, but apart from a conceptual similarity to cortical structure they stray away from their biological archetypes.
Their recent success has prompted many efforts to implement them with more biologically plausible components, but the computational advantages thereof have so far proven elusive.
In this work, we focus on two well-established biological facts: spike-based communication between neurons and a limited pool of synaptic resources (neurotransmitters).
We argue that, in combination, these two mechanisms can endow networks with computational capabilities that are otherwise difficult to achieve.
In particular, in the context of probabilistic inference, we show how plastic synapses bolster the generative capabilities of spiking networks while requiring only a small, local computational overhead, as opposed to the classical tempering solutions for their conventional counterparts.
Our work thereby highlights important computational consequences of biological features that might otherwise appear as mere engineering limitations or artifacts of evolution.
\section*{Abstract}
Spiking networks that perform probabilistic inference have been proposed both as models of cortical computation and as candidates for solving problems in machine learning.
However, the evidence for spike-based computation being in any way superior to non-spiking alternatives remains scarce.
We propose that short-term plasticity can provide spiking networks with distinct computational advantages compared to their classical counterparts.
In this work, we use networks of leaky integrate-and-fire neurons that are trained to perform both discriminative and generative tasks in their forward and backward information processing paths, respectively.
During training, the energy landscape associated with their dynamics becomes highly diverse, with deep attractor basins separated by high barriers.
Classical algorithms solve this problem by employing various tempering techniques, which are both computationally demanding and require global state updates.
We demonstrate how similar results can be achieved in spiking networks endowed with local short-term synaptic plasticity.
Additionally, we discuss how these networks can even outperform tempering-based approaches when the training data is imbalanced.
We thereby show how biologically inspired, local, spike-triggered synaptic dynamics based simply on a limited pool of synaptic resources can allow spiking networks to outperform their non-spiking relatives.
\section*{Introduction}
Neural networks are, once again, in the focus of both the artificial and the biological intelligence communities. 
Originally inspired by the dynamics and architecture of cortical networks \cite{mcculloch1943logical, rosenblatt1958perceptron}, they have increasingly strayed away from their biological archetypes, prompting questions about their relevance for understanding the brain \cite{crick1989recent,stork1989backpropagation}.
However, their recent hardware-fueled dominance \cite{lecun2015deep} has motivated renewed efforts to align them with biologically more plausible models \cite{lillicrap2016random, lee2016training, neftci2017event, petrovici2017pattern}.
Moreover, neural networks have been used to explain some aspects of in-vivo cortical dynamics \cite{zipser1988back, kriegeskorte2015deep}. 

Two questions are immanent to these efforts: From a machine learning perspective, how useful are spike-based versions of deep neural networks? And from a biological perspective, how much can we learn about the brain from artificial neural networks?
Much of the recent work on neural networks has focused on the "forward" computation pathway, i.e., learning pattern classification through error backpropagation \cite{schmidhuber2015deep}.
However, the "backward" pathway required for generative models has also made significant progress \cite{hinton2012deep, goodfellow2014generative}. 
A key aspect to the success of a generative model is its capability to mix, i.e., travel through the probability landscape that it needs to represent.
The performace gain of recently proposed models is to a large extent due to refined mixing algorithms, most of which are based on a form of simulated tempering \cite{desjardins2010parallel, salakhutdinov2010learning, bengio2013better}. 

The discriminative capacity of the neocortex is well-established, as evidenced by the difficulty of artificial systems to achieve superhuman classification performance \cite{schmidhuber2015deep}. 
Simultaneously however, the brain also appears to learn a generative model of its sensory environment \cite{fiser2010statistically, jezek2011theta, hindy2016linking}. 
How these capabilities are achieved remains an open question, but it is unlikely that complex tempering schedules are at work. 

One mechanism that is capable of modulating synaptic weights and thereby shaping the probability landscape of a neural network is short-term synaptic plasticity. 
In this work, we investigate the ability of this biologically ubiquitous mechanism to improve the mixing capabilities of generative neural networks. 
Furthermore, we show how hierarchical spiking networks endowed with short-term plasticity can simultaneously become good discriminative and generative models, a feature that is difficult to achieve due to the conflicting nature of these two tasks. 
We thereby offer a potential explanation for the generative capabilities of cortical networks, while at the same time proposing a simple but efficient mechanism to bolster the usefulness of spiking networks for machine learning applications. 
This can be of particular interest in combination with spiking neuromorphic systems which, compared to conventional simulation platforms, implement fast and energy-efficient physical models of neuro-synaptic dynamics \cite{pfeil2013six, schemmel2010wafer}. 
\section*{Methods}

    \begin{figure}[t]
        \centering
        \includegraphics[width=1.\linewidth]{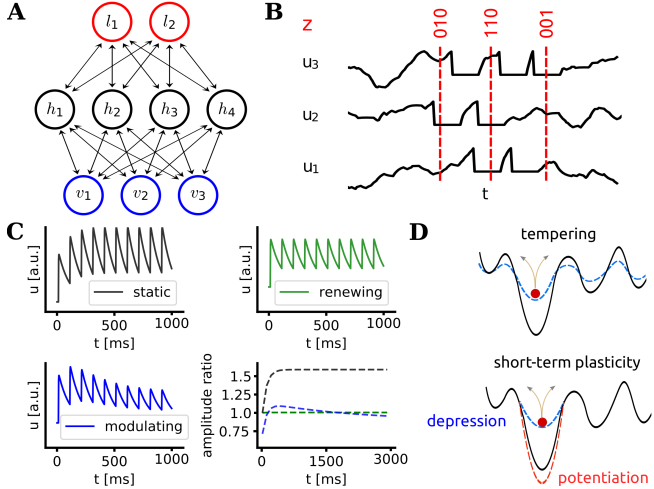}
        \caption{
            \tb{(A)} Structure of a hierarchical sampling spiking network.
                     Its classical counterpart is a restricted Boltzmann machine with a visible ($v$), hidden ($h$) and label ($l$) layer.
            \tb{(B)} Interpretation of states as samples in a spiking network.
                     A neuron with a freely evolving membrane potential is said to be in the state $z_k=0$ and switches to the state $z_k=1$ upon firing, where it stays for the duration of the refractory period.
            \tb{(C)} Sketch of the membrane potential evolution for three relevant scenarios: static (black), renewing (green) and modulated (blue) synapses.
                     Bottom right: envelope of the PSP height for three parameter sets $(U_0, \taurec)$ from the manuscript: $(1, 0)$ (black), $(1, \tausyn)$ (green) and $(0.01, \SI{280}{ms})$ (blue).
                     Note how the latter only weakly modulates the PSP height.
            \tb{(D)} In order to correctly sample from a posterior distribution, a network needs to be able to mix, i.e., traverse barriers between low-energy basins.
                     To facilitate mixing, tempering methods globally rescale the energy landscape with an inverse temperature (top).
                     In contrast, STP can be viewed as only modulating the energy landscape locally, thereby only affecting the currently active attractor (bottom).
        }
        \label{fig:f1}
    \end{figure}

    We start with a brief introduction of Boltzmann machines as generative models and their spike-based implementation.
    We then describe the problem of mixing and outline the essential elements of tempering-based solutions.
    Finally, we discuss the model of short-term plasticity that we later use in our spiking networks.
    Supplementary information (SI) available in the last page.

    \subsection*{Boltzmann machines and spiking networks}

        Among the neural networks proposed as generative models for high-dimensional input, Boltzmann machines (BMs) \cite{smolensky1986information} are arguably the most prominent \cite{larochelle2008classification, salakhutdinov2009deep, dahl2010phone, srivastava2012multimodal}.
        Neurons in BMs are binary units with states $z_k \in \left\{0,1\right\}$.
        These states are typically updated in a sequential schedule in a way that implements Gibbs sampling from a target Boltzmann distribution
        \begin{align}
            \textstyle \pB\left(\bs z \middle| \bs W, \bs b\right) = Z^{-1} \exp [-\beta E(\bs z)]
        \end{align}
        with the inverse temperature $\beta \in (0, 1]$, partition function $Z$ and the energy function $E(\bs z) = -\bs z^\T \bs W \bs z/2 - \bs z^\T \bs b$ parametrized by the weight matrix $\bs W$ and bias vector $\bs b$.
        This is achieved by having each neuron compute a local "membrane potential" as the log-odds of its conditional firing probability, which for the Boltzmann distribution is equal to a weighted sum over input activities:
        \begin{align}
            \textstyle u_k = \ln \frac{\pB\left(z_k=1 \middle| z_{\non k}\right)}{\pB\left(z_k=0 \middle| z_{\non k}\right)} = \sum_{i \neq k} w_{ki} z_i + b_k \quad .
            \label{eqn:lcc}
        \end{align}
        Consequently, state updates are computed using a logistic activation function $p(z_k=1) = \left[ 1+\exp(u_k) \right]^{-1} =: \sigma(u_k)$.
        
        In a restricted Boltzmann machine (RBM), the units are subdivided into a visible and a hidden layer, with no within-layer connections (Fig.~\ref{fig:f1}A).
        To enable classification, an additional label layer can be added, which for training purposes can be treated as part of the visible layer.
        During training, weights and biases are iteratively updated in order to optimize the marginal distribution $p(\bs v, \bs l | \bs h)$ as the underlying distribution for the set of training samples.
        
        Recently, it has been shown how networks of spiking neurons can perform equivalent computations \cite{petrovici2016stochastic}, which we briefly outline in the following.
        The building blocks for our spiking networks are conductance-based leaky integrate-and-fire (LIF) neurons, with membrane potential dynamics governed by
        \begin{align}
            \textstyle \Cm \frac{\mathrm d u}{\mathrm d t} = \gl (\El - u) + \Isyn \quad ,
        \end{align}
        where $\Cm$ is the membrane capacitance, $\gl$ and $\El$ the leak conductance and potential, and $\Isyn$ the synaptic current.
        Neurons fire upon reaching a threshold voltage, which causes the membrane to be clamped to a reset potential for a duration equal to the refractory period $\tauref$.
        The synaptic current is modeled as a sum of exponential kernels triggered by presynaptic spikes $s$ with a synaptic time constant $\tausyn$ and weighted by synaptic weights $W_i(t)$ and reversal potentials $\Erev_i$:
        \begin{align}
            \textstyle \Isyn(t) = \sum_s \sum_i W_i(t) \left( \Erev_i - u \right) \exp\left[ -(t-t^s)/\tausyn \right] \enskip .
            \label{eqn:isyn}
        \end{align}
        The temporal dependence of the synaptic weights accounts for the STP mechanism we discuss later.
        
        Each neuron receives both functional synaptic input from other neurons within the network and diffuse background input from external neurons that can be modeled as Poisson spike trains.
        The latter causes the neuron to fire stochastically.
        Since, at the level of spikes, the output of a neuron can be considered binary, we associate a binary random variable $z_k$ to each neuron.
        As a neuron never fires within the refractory period, it is natural to set $z_k=1$ for $t \in [t_k^s, t_k^s+\tauref)$ and 0 otherwise (Fig.~\ref{fig:f1}B).
        
        For constant functional synaptic input, the mean firing rate of such a neuron is proportional to its activation function $p(z_k\!\!=\!\!1)$.
        By applying strong background input, we lift neurons into a high-conductance state \cite[HCS,][]{destexhe2003hcs}, which molds their activation function into an approximately logistic shape \cite{petrovici2015high}:
        \begin{align}
            \textstyle p(z_k=1) \approx \sigma(\alpha \uf_k - \beta) \quad ,
            \label{eqn:calib}
        \end{align}
        with scaling parameters $\alpha$ and $\beta$, where $\uf_k$ represents the functional, i.e., background-free, membrane potential.
        Similarly to Gibbs sampling, the functional membrane potential thereby fulfills the local computability condition (Eqn.~\ref{eqn:lcc}), which is a sufficient computational prerequisite for sampling in neural networks \cite{smolensky1986information,buesing2011neural}.
        The scaling parameters can be derived analytically and allow a direct translation of the BM parameters $\bs W$ and $\bs b$ to the corresponding parameters in the biological domain (SI, Sec.~1).

    \subsection*{Tempering vs. short-term plasticity}
        
        When trained from data, the energy landscape $E(\bs z)$ is shaped in a way that assigns low energy values (modes) to the samples in the training data.
        If this dataset is composed of very dissimilar classes, training algorithms tend to separate them by high energy barriers.
        As their height grows during training, Gibbs sampling becomes increasingly ineffective at covering the entire relevant state space, as reflected by a high correlation between consecutive samples caused by the component-wise update of states.
        Consequently, a BM would need longer to converge towards its underlying distribution.
        This problem becomes particularly inconvenient when dealing with complex, real-world data, or when an agent must rely on the prediction of the network to make a fast decision.
        
        The ability of a sampling-based generative model to jump across energy barriers, also known as mixing, has therefore received significant attention \cite{marinari1992simulated, wang2001efficient, salakhutdinov2010learning, bengio2013better}.
        Many of these methods rely on some version of simulated tempering, which modifies the temperature parameter $\beta_T$ in order to globally flatten the network's energy landscape (Fig.~\ref{fig:f1}D).
        Therefore, in addition to conventional Gibbs sampling, we use the adaptive simulated tempering algorithm \cite[AST][]{salakhutdinov2010learning} as a benchmark for our spiking networks (SI, Sec.~2).
        
        While greatly increasing the mixing capabilities of generative networks, it is important to note that all tempering schedules come with a cost of their own, both because they require additional computations and because they only gather valid samples at low temperatures ($\beta \approx 1$), thereby effectively slowing down the sampling process.
        Furthermore, they require parameter changes that assume knowledge about the global state of the network, which is difficult to reconcile with biology.
        This motivates the search for a local update rule that has biological relevance, improves mixing and can be embedded in spiking networks.
        
        In biological neural networks, the momentary synaptic interaction strength is reflected in the size of the elicited postsynaptic potential (PSP).
        In dynamic synapses, this value may change over time depending on the presynaptic activity.
        To model this dependence, we use the Tsodyks-Markram model of short-term plasticity \cite[STP,][]{tsodyks1998neural}:
        \begin{align}
            \textstyle \PSP &\propto \textstyle w \cdot U \cdot R \label{eqn:tso1} \\
            \textstyle \mathrm{d} R / \mathrm{d} t &= \textstyle (1-R)/\taurec - U \cdot R \cdot \delta(t-t_s) \label{eqn:tso2} \\
            \textstyle \mathrm{d} U / \mathrm{d} t &= \textstyle -U/\taufac + U_0 \cdot (1-U) \cdot \delta(t-t_s) \label{eqn:tso3} \quad .
        \end{align}
        Here, $w$ represents the (static) synaptic weight and $U \in [0,1]$ the utilized fraction of available synaptic resources $R \in [0,1]$.
        Upon arrival of a presynaptic spike at time $t_s$, the synapse is depressed by subtracting $U$ from $R$, which recovers exponentially with the time constant $\taurec$.
        Facilitation is modeled by a simultaneous increase in $U$, followed by an exponential decay with time constant $\taufac$.
        
        Since both tempering and STP effectively modify the energy landscape by changing network parameters during sampling, they clearly bear some conceptual resemblance.
        However, while tempering simultaneously affects all synaptic weights, STP only affects the efferent connections of those neurons that are simultaneously active at a given moment in time.
        Therefore, in contrast to the global modifications of the energy landscape incurred by tempering, STP has a more local effect, as sketched in Fig.~\ref{fig:f1}D.
        Note that this effect is not equivalent to neuronal adaptation, because it does not prohibit neurons from remaining active over extended periods, which is essential for generating consecutive patterns with significant overlap.

\section*{Results}

    We study the effects of STP on the performance of spiking networks trained for different tasks.
    We start by discussing how STP can improve the sampling accuracy of small networks configured to sample from a fully specified target distribution, even when the energy landscape is shallow enough to not cause mixing problems.
    This is no longer the case for hierarchical networks trained directly on data, for which we study the influence of STP on both their generative and their discriminative properties.
    Furthermore, we show how STP can aid pattern completion in a network trained on a highly imbalanced dataset.
    For all spiking network simulations, we used NEST \cite{diesmann2001nest} with PyNN \cite{davison2008pynn} as a front-end.
      
    \subsection*{Sampling from a fully specified target distribution}
    
        By modulating synaptic interactions, STP shapes the sampled distribution.
        This can be helpful when a spiking network needs to approximate a distribution that is otherwise incompatible with biological neuro-synaptic dynamics, as we discuss in the following.
        
        Consider the case where the target distribution of the spiking network is a Boltzmann distribution.
        When a neuron needs to continuously represent a state $z_k(t)=1$ for an extended period, it fires a sequence of $n$ spikes at maximum frequency $1/\tauref$.
        Following Eqn.~\ref{eqn:lcc}, the resulting PSPs should increase a postsynaptic neuron's membrane by a constant $\Delta u_i = w_{ik}$, which implies a rectangular PSP shape.
        However, this is not a realistic shape for a more biologically plausible scenario, where PSPs have an exponentially shaped decay.
        This causes them to accumulate (Fig.~\ref{fig:f1}C), such that the average increment $\expect{\Delta u_i}_n$ becomes a function of the burst length $n$, thereby distorting the sampled distribution.
        
\begin{figure}
            \centering
            \includegraphics[width=1.\linewidth]{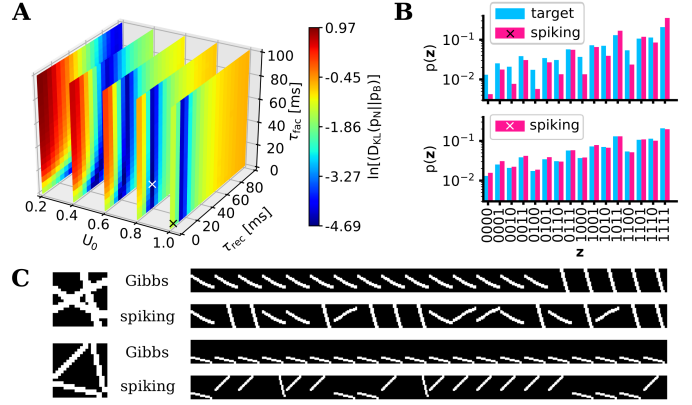}
            \caption{
                \tb{(A)} Kullback-Leibler divergence between sampled ($p_\mathrm{N}$) and target ($p_\mathrm{B}$) distribution of a spiking network with 10 neurons (5 hidden, 5 visible) for different STP parameters $(U_0,\taurec,\taufac)$.
                         Note that many different parameter combinations lead to close to optimal (white cross) sampling, but static synapses (black cross) are not among them.
                \tb{(B)} Distribution sampled by the spiking network for two different configurations of synaptic parameters.
                         Depressing synapses (bottom) allow the network to come much closer to the target distribution (blue) than non-plastic ones (top).
                \tb{(C)} Left: Training data for the easy (top) and hard (bottom) learning scenario (individual images are overlapped).
                         Right: Sequence of images generated by a Gibbs sampler and an STP-endowed spiking network with equivalent parameters $\bs W$ and $\bs b$.
                }
            \label{fig:f2}
        \end{figure}

        Synaptic depression can mitigate this effect (Fig.~\ref{fig:f2}B) by causing a gradual decrease in the amplitude of consecutive PSPs.
        Indeed, when sweeping over the $(U_0,\taurec,\taufac)$ parameter space (Fig.~\ref{fig:f2}A), we find that an optimal reproduction of the target distribution is achieved for $\taurec\approx\SI{15}{ms}$, which is close to the synaptic time constant of $\tausyn=\SI{10}{ms}$.
        This affords an intuitive explanation: In the HCS, the effective membrane time constant becomes small and $\tausyn$ dominates the PSP decay.
        If the recovery of synaptic resources $R$ (Eqn.~\ref{eqn:tso2}) happens at the same speed as the PSP decay, the STP mechanism essentially emulates a renewing synapse with an approximately constant running average (Fig.~\ref{fig:f1}C).
        The slightly larger optimal recovery time constant further compensates for the long tails of exponential PSPs, which potentiate interaction strengths compared to the ideal case of rectangular PSPs (SI, Sec.1).
        Note that the manifold for which the target distribution is close-to-optimally reproduced contains many different STP configurations, including the range of biologically observed parameters \cite{wang2006heterogeneity, costa2013probabilistic}, but not the $(u,\taurec,\taufac)=(1,0,0)$ triplet for static synapses (Fig.~\ref{fig:f2}A).
        
        For this example, we used a fully specified target distribution $\pB\left(\bs z \middle| \bs W, \bs b\right)$; training was not needed, as synaptic weights can be computed directly from the parameters $\bs W$ and $\bs b$ (SI, Sec.1).
        Here, we used a target Boltzmann distribution with randomly drawn parameters that produce a diverse energy landscape, but not so rough as to create problems with mixing.
        This changes when the network parameters are learned from data, as we discuss in the following.
    
    \subsection*{Mixing in a simple learning scenario}
    
        Borrowing from observations in the early visual system, we generated images of oriented bars.
        The bars were positioned in a way that gave rise to an "easy" (overlapping) and a "hard" (non-overlapping) dataset (Fig.~\ref{fig:f2}C).
        We then trained a two-layer hierarchical network (400 visible, 30 hidden units) on each of these datasets using a version of the wake-sleep algorithm \cite{salakhutdinov2010learning} (SI, Sec.~2).
        Intuitively, the difficulty of learning a generative model of this data increases when the bars have little or no overlap: in this case, training gives rise to three nearly disjoint populations that have, on average, excitatory connections within and inhibitory connections between them.
        The emergence of such a population-based winner-take-all structure can be characterized by the mean interaction strength $\bar w_{ij} = \expect{\bs z_i^\T} \bs W \expect{\bs z_j}$ between two population activity vectors $\expect{\bs z_i}$ and $\expect{\bs z_j}$, which represent the average network activity during the presentation of the $i$th and $j$th input pattern, respectively.
        For the easy dataset, learning gave rise to a mean within-population interaction strength of $\expect{\bar w_{ii}}_i = 92.75$ and a mean between-population interaction strength of $\expect{\bar w_{ij}}_{i \neq j} = -145.48$.
        These values changed to $\expect{\bar w_{ii}}_i = 102.82$ and $\expect{\bar w_{ij}}_{i \neq j} = -164.66$ for the hard dataset, reflecting the increased competition and disjointedness between the three emerging populations.
        STP, however, can weaken active synapses, temporarily reducing $\left| \expect{\bar w} \right|$ to enable switching between attractors.
        
        The learned parameter set was used to compare the performance of classical Gibbs sampling and STP-endowed spiking networks (Fig.~\ref{fig:f2}C).
        For the easy dataset, both the Gibbs sampler and the spiking network were able to mix, although the former spent on average 100 times longer in the same mode before switching, thereby requiring more time to converge to the target distribution.
        For the hard dataset, the spiking networks retained their ability to mix, whereas Gibbs sampling was unable to leave the (randomly initialized) local mode.
        These observations mirror those found in studies of cortical attractor networks \cite{lundqvist2006attractor}.
        While this simple experimental setup was specifically designed to illustrate the potential problems of sampling-based generative models and the ability of STP-endowed spiking networks to circumvent them, we show in the following that these properties are preserved in more complex scenarios.
    
    \subsection*{Generation and classification of handwritten digits}
    
        \begin{figure*}[t]
        \centering
        \includegraphics[width=1.\linewidth]{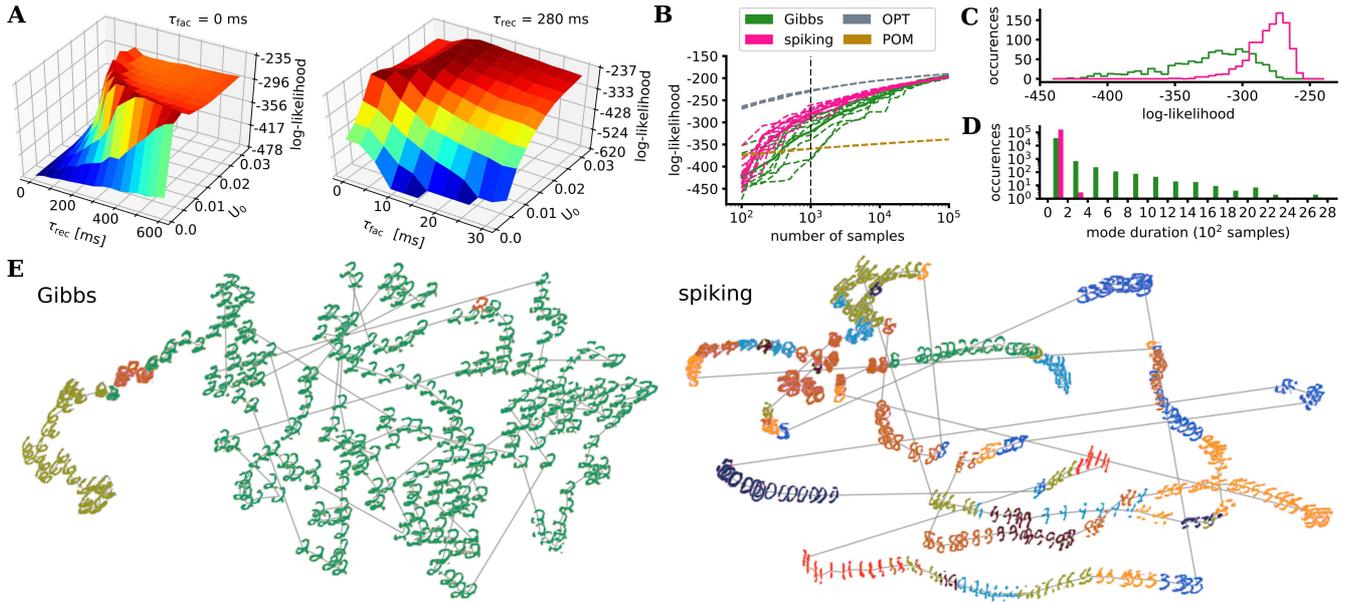}
        \caption{
            Superior generative performance of an STP-endowed spiking network compared to an equivalent Gibbs sampler.
            (A) 2D parameter scans of the STP parameters $(U_0,\taurec,\taufac)$ with multiple configurations leading to good generative performance.
            (B) Log-likelihood (ISL) of the test set calculated from an increasing number of samples.
                Each sampling method was simulated with 10 different random seeds.
                The ISLs of an optimal sampler with the same parameters (OPT, gray) and the product of marginals (POM, brown) are shown for comparison (see SI Sec.~3).
            (C) Direct comparison between the two sampling methods for $10^3$ samples, equivalent to a sampling duration of \SI{10}{s} in the biological domain.
                ISL histogram generated from 100 random seeds.
            (D) Histogram of times spent within the same mode when no visible units are clamped.
            (E) tSNE plots of images produced by the two methods over 1800 consecutive samples.
                For every 6th of these samples, an output image is shown.
                Consecutive images are connected by gray lines.
                Different colors represent different image classes, defined by the label unit that showed the highest activity at the time the sample was generated.
                Note that tSNE inherently normalizes the area of the 2D projection; the volume of phase space covered by the Gibbs chain is, in fact, much smaller than the one covered by the spiking network.
            }
        \label{fig:f3}
        \end{figure*}
    
        The problem of mixing becomes even more pronounced when dealing with larger, more complex datasets.
        Here, we trained a hierarchical 3-layer network with 784 visible, 600 hidden and 10 label units on handwritten digits from the MNIST dataset \cite{lecun1998mnist}.
        By treating the label units as part of the visible layer during training, we simultaneously trained a generative and a discriminative model of the data.
        This objective is particularly challenging, because mechanisms that improve mixing tend to disrupt classification and vice-versa.
        
        To evaluate the quality of generated samples, we computed a log-likelihood estimation of 2000 test images (not used during training) using the indirect sampling likelihood (ISL) method \cite[][see also SI]{breuleux2010unlearning, desjardins2010tempered}.
        Due to the size of the network, a full scan of the parameter space for finding optimal STP parameters was no longer feasible.
        Therefore, starting from a good parameter set found by trial and error, we performed two 2D-scans of the $(U_0,\taurec,\taufac)$ parameter space (Fig.~\ref{fig:f3}A).
        As in the previous examples, we found short-term depression to be essential for achieving high ISL values.
        Furthermore, a small value of $U_0$ combined with short-term facilitation was also beneficial, allowing an initial strengthening followed by a weakening of the active attractor, as sketched in Fig.~\ref{fig:f1}C,D.
        Similar observations have been made in cortex, where STP can promote the enhancement of transients \cite{abbott2004synaptic}.
        
        We used one of the optimal STP parameter sets $(U_0=0.01, \taurec=\SI{280}{ms})$ to compare the generative performance of spiking networks to classical Gibbs sampling.
        Due to its improved mixing capability, the spiking network was able to quickly cover a large portion of the relevant state space, as reflected by a faster ISL gain during sampling (Fig.~\ref{fig:f3}B).
        This is a systematic effect and only weakly dependent on initial conditions, as can be seen in Fig.~\ref{fig:f3}C, which shows a histogram over 100 random seeds.
        For this comparison, we chose a sampling duration of \SI{10}{s} as a conservative estimate for the maximum duration for a biological agent to experience stable stimulus conditions and therefore sample from a stable target distribution.
        The faster mixing is the result of the spiking network's ability to jump out of local attractors, which is reflected in a much shorter time spent on average within the same mode (Fig.~\ref{fig:f3}D).
        Here, we defined a mode as the dominant class of the currently represented image; a mode was therefore defined by the identity of the neuron in the label layer with the highest firing rate.
        
        It is important to note that, due to the STP-modulated interaction, the spiking network does not sample from the exact same distribution as the Gibbs sampler, despite using an equivalent $(\bs W, \bs b)$ parameter set.
        However, for a very large number of samples ($>10^5$), the two methods converge towards the same ISL (Fig.~\ref{fig:f3}B), indicating that the discrepancy in performance for shorter sampling durations is not due to a fundamental difference in their respective ground truths.
        
        While the ISL, as an abstract quantity, provides a useful numerical gauge of the quality of a generative model, a direct depiction of the produced images is particularly instructive.
        Here, we used the t-distributed stochastic neighbor embedding (tSNE) method \cite[][see also SI]{maaten2008visualizing} for a 2D-embedding of the high-dimensional sampled distribution.
        The similarity between samples is largely reflected by their 2D distance and a large jump can be interpreted as a switch between attractors.
        As seen in Fig.~\ref{fig:f3}E, the spiking network produces a significantly more diverse set of samples compared to the Gibbs sampler.
        
        When the visible layer is clamped to a particular input, the same network can be used as a discriminative model of the learned data.
        Using the same parameters as for the generative task, the benchmark Gibbs sampler obtained a classification accuracy of 93.4\% on the MNIST test data.
        The spiking network with STP performed only slightly worse, at 93.2\%.
        The additional generative capabilites gained by the spiking networks through STP were therefore not strongly detrimental to their classification accuracy.
    
    \subsection*{Modeling an imbalanced dataset}

        \begin{figure}[t]
            \centering
            \includegraphics[width=1.\linewidth]{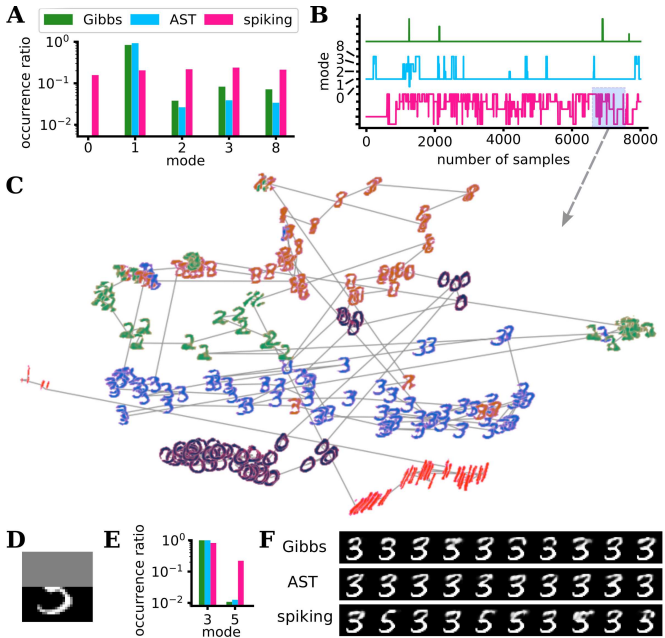}
            \caption{
                Comparison of Gibbs and AST samplers with STP-endowed spiking networks for imbalanced training data.
                (A) Histogram of relative time spent in different modes calculated from 16,000 samples.
                (B) Mode evolution over 8,000 consecutive samples.
                (C) tSNE plot of images generated by the spiking network over a duration of \SI{10}{s}, with \SI{40}{ms} between consecutive images.
                (D) Ambiguous input to the visible layer.
                    The upper half is not clamped and free to complete the pattern.
                (E) Histogram of relative time spent in different modes during the pattern completion task, measured over 20,000 consecutive samples.
                (F) Comparison of the sequence of images generated by the different methods over 5000 samples (only every 500th is shown).
                }
            \label{fig:f4}
        \end{figure}

        In many real-world scenarios, the available data is imbalanced, with much of the data belonging to one class and significantly less samples being distributed over others.
        It is well-known that imbalanced data can cause severe problems for data mining and classification \cite{chawla2005data, garcia2009evolutionary}.
        One solution is to create a more balanced dataset from the imbalanced one, which can be achieved by methods such as under- or over-sampling \cite{garcia2009evolutionary, chawla2002smote}.
        However, such an a-priori modification of the input data does not seem biologically plausible.
        Still, cognitive biological agents appear to easily overcome this problem: humans will have little difficulty imagining a platypus from seeing only its bill, despite having likely seen many more ducks throughout their lifetime.
        Spiking networks with STP provide a simple solution to the problem of imbalanced training data, without any need for preprocessing.
        
        We generated an imbalanced dataset of 1000 images by randomly selecting 820 digits of class "1" and 45 from the "0", "2", "3" and "8" classes.
        After training, we compared the generative output of a Gibbs sampler, an AST sampler and a spiking network with STP.
        Note that the effective sampling speed of AST is roughly 20 times slower compared to Gibbs sampling, since most of the produced samples are not considered valid.
        In this scenario, it becomes particularly useful that the spiking network transiently modifies the learned data distribution (Fig.~\ref{fig:f4}A).
        The STP-induced weakening of active attractors balances out their activity, thereby negating the inherent imbalance induced by the training data.
        Furthermore, as observed before, the spiking network switches faster between modes (Fig.~\ref{fig:f4}B,C).
        
        These abilities become particularly useful in a scenario of inference based on incomplete information, for which pattern completion is a prime example.
        Here, we used a training set with 6 majority classes ("0", "1", "2", "3", "4", "6", 800 samples each) and one minority class (200 samples of "5").
        We generated an ambigous image by clamping the lower half of the visible layer to a configuration compatible with both a "3" and a "5" (Fig.~\ref{fig:f4}D).
        While Gibbs and AST strongly undersample the minority class, the spiking network produces a much more balanced set of images, with swift transitions between modes (Fig.~\ref{fig:f4}E,F).
        The estimate of the possible realities underlying the incomplete observation is therefore improved both on long and on short time scales.
        This can be particularly useful for an agent in need of a quick reaction, as, for example, often required in nature in a fight-or-flight scenario.

\section*{Discussion}
We have shown how a combination of event-based communication and short-term plasticity can enhance the ability of neural networks to perform probabilistic inference in high-dimensional data spaces.
Here, a spike-triggered plasticity rule played a similar role to simulated tempering methods used for classical neural networks, but without requiring complex computations on the global network state or long waiting times between valid samples.
The spiking networks outperformed their classical counterparts as generative models of real-world data, with little disturbance to their classification capability, which we expect to be largely remediable by additional fine-tuning of the network parameters.
Furthermore, they were also able to cope with imbalanced training data, as demonstrated by their superior performance in a pattern completion task on ambiguous input. 
Intriguingly, the synaptic parameters used to achieve this performance are compatible to experimental data \cite{wang2006heterogeneity,  costa2013probabilistic}. \\

In a physical system such as a biological brain, the studied plasticity mechanism essentially comes for free, as it only requires a limited pool of synaptic resources.
Together with other activity-modulating mechanisms such as neuronal adaptation, it could be a key contributor to the ability of the brain to navigate efficiently in a very-high-dimensional stimulus space.
Importantly, these networks provide immediate computational advantages for spike-based neuromorphic devices, facilitating the development of efficient artificial agents that replicate the inferential capabilities of their biological archetypes.
\section*{Acknowledgments}
We thank Johannes Bill for valuable discussions and comments.
This research was supported by EU grants \#269921 (BrainScaleS), \#604102 and \#720270 (Human Brain Project), the Heidelberg Graduate School of Fundamental Physics and the Manfred St\"ark Foundation.

\clearpage

\section*{Supplementary Information (SI)}

    \subsection*{1. Spiking networks}
    
\begin{figure}[h]
            \centering
            \includegraphics[width=1.\linewidth]{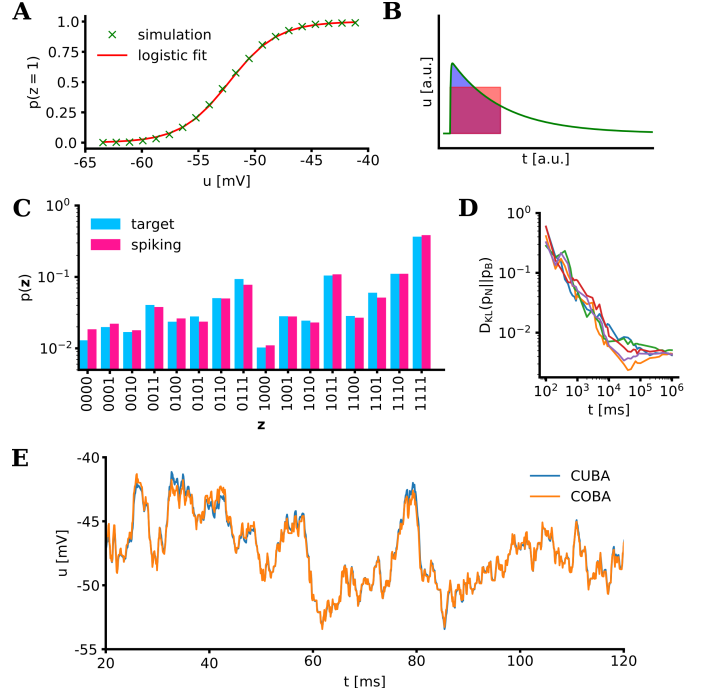}
            \caption{
                (A) Activation function of an LIF neuron in the HCS and logistic fit.
                (B) Sketch of synaptic weight translation (Eqn.~\ref{eq:weighttranslation}).
                (C) Sampled distribution of a fully connected 4-neuron LIF network vs. target distribution.
                (D) Evolution of Kullback-Leibler divergence between sampled ($p_\mathrm{N}$) and target ($p_\mathrm{B}$) distribution for 5 different random seeds.
                (E) Free membrane potential ($\vartheta=0$) of a biologically plausible COBA LIF neuron in the HCS compared to an equivalent CUBA LIF neuron (parameters given in Tab.~\ref{tab:neuronparams}).
                }
                \label{fig:sif1}
        \end{figure}

        To generate our spiking sampling networks, we follow \cite{petrovici2016stochastic}.
        We emulate an HCS by stimulating the LIF neurons with balanced excitatory and inhibitory Poisson noise 
        This produces an approximately logistic activation function (Fig.\ref{fig:sif1}A), parametrized by a shift $\beta$ and a scaling parameter $\alpha$ (Eqn.~\ref{eqn:calib}).
        These parameters can be used to translate synaptic interaction strengths from the Boltzmann domain to synaptic conductances:
        \begin{align}
            W_{kj} & = \frac{1}{\alpha \Cm} \frac{w_{kj} \left(\Erev_{kj} - \mu\right)}{1-\frac{\tausyn}{\taueff}} \nonumber \\
                   & \left[\tausyn(e^{-1} - 1) - \taueff \left( e^{- \frac{\tausyn}{\taueff}} - 1 \right) \right] \quad ,
            \label{eq:weighttranslation}
        \end{align}
        where $W_{kj}$ denotes the peak synaptic conductance (see Eqn.~\ref{eqn:isyn}), $\Cm$ the membrane capacitance, $w_{kj}$ the abstract Boltzmann weight, $\Erev_{kj}$ the corresponding reversal potential, $\mu$ the mean free membrane potential, $\tausyn$ the synaptic time constant and $\taueff=\Cm/\expect{\gtot}$ the (mean) effective membrane time constant.
        This corresponds to a match of the average interaction during the refractory period of the presynaptic neuron (Fig.\ref{fig:sif1}B).
        This setup allows an accurate sampling from target Boltzmann distributions (Fig.\ref{fig:sif1}C,D).
        
        To speed up simulations, we used an effective current-based (CUBA) model to replace the conductance-based (COBA) one.
        Fig.~\ref{fig:sif1}E shows a comparison between the two models.
        Under appropriate parametrization, we could reduce the background input rates from $\nu=\SI{5}{kHz}$ to $\nu=\SI{0.4}{kHz}$.
        
        \begin{table}[h]
            \centering
            \caption{Neuron parameters}
            \begin{tabular}{cccl}
                \hline
                 & COBA & CUBA & \\
                \hline
                $\Cm$ & \SI{0.1}{nF} & \SI{0.2}{nF} & membrane capacitance \\
                $\taum$ & \SI{20}{ms} & \SI{0.1}{ms} &  membrane time constant \\
                $\tauref$ & \SI{10}{ms} & \SI{10}{ms} & refractory time constant \\
                $\tausyn$ & \SI{10}{ms} & \SI{10}{ms} &  synaptic time constant \\
                $\uthr$ & \SI{-50}{mV} & \SI{-50}{mV} &  threshold voltage \\
                $\rho$ & \SI{-53}{mV} & \SI{-50.01}{mV} & reset potential \\
                $\Ereve$ & \SI{0}{mV} & - & excitatory reversal potential \\
                $\Erevi$ & \SI{100}{mV} & - & inhibitory reversal potential \\
                \hline
            \end{tabular}
            \label{tab:neuronparams} 
        \end{table}
        
    \subsection*{2. Training}
    
        To speed up training, we used RBMs with binary units, followed by a mapping of the resulting parameters to the spiking-network domain as described above.
        As a learning algorithm, we used the coupled adaptive simulated tempering (CAST) method \cite{salakhutdinov2010learning}.
        In CAST, two instances of the RBM are simulated in parallel, with one of them staying at a constant inverse temperature $\beta=1$ and the other one using AST for mixing.
        In AST, states $\bs z^{(t+1)}$ are updated by Gibbs sampling from $p(\bs z | \beta_T^{(t)})$.
        After each state update, the temperature is itself updated by an adaptive rule that ensures the algorithm spends a roughly equal amount of time at each value $\beta_T$ (Tab.~\ref{tab:astalgo}).
         \begin{table}[h]
             \centering
             \caption{Adaptive simulated tempering}
             \setlength\tabcolsep{2.5pt} 
             \begin{tabular}{ll}               
                 \hline \\
                 1: & Given adaptive weights $\{\mathbf{g}_k\}^K_{k=1}$ and the initial configuration of \\
                    & the state $\bs{z}^1$ at temperature 1, $k$ = 1: \\
                 2: & \textbf{for} $t = 1:T$ (number of iterations) \textbf{do}\\
                 3: & \quad Given $\bs{z}^t$, sample a new state $\bs{z}^{t+1}$ from $p(\bs{z}|k^t)$ \\
                    & \quad by Gibbs sampling.\\
                 4: & \quad Given $k^t$, sample $k^{t+1}$ from proposal distribution $q(k^{t+1} \leftarrow k^t)$. \\
                    & \quad Accept with probability: $\mathrm{min} \left(  1, \frac{p(\bs{z}^{t+1}, k^{t+1})q(k^t \leftarrow k^{t+1}) g_{k^t}}{p(\bs{z}^{t+1}, k^n)q(k^{t+1} \leftarrow k^t) g_{k^{t+1}}}  \right)$ \\
                 5: & \quad Update adaptive adjusting factors: \\
                    & \quad $g_i^{t+1} = g_i^t (1+ \gamma_t  I(k^{t+1} = i )), \; i = 1,...,K.  \nonumber $\\
                 6: & \textbf{end for}\\
                 7: & Collect data: Obtain (dependent) samples from target distribution \\
                    & $p(z)$ by keeping $k=1$. \\ \\
                 \hline
             \end{tabular}
             \label{tab:astalgo} 
         \end{table}   
        
        The used hyperparameters (number of epochs $T$, batch size $N$, learning rate $\eta$) were based on suggestions from previous work \cite{hinton2010practical} and empirical experience.
        For all datasets, we used 20 equidistant inverse temperatures $\beta_k \in [0.9, 1]$. The adaptive weights $\{\mathbf{g}_k\}^K_{k=1}$ were initialized to 1 for all temperatures and as $\gamma_t \rightarrow 0$ 
        the adaptive weights will converge. In all experiments, we set $\gamma_t$ as $90/(150+t)$. 
        For the bar example (Fig.~\ref{fig:f2}), we used $T=100,000$, $N=3$ and $\eta=10/(2000+t)$.
        For the full MNIST example (Fig.~\ref{fig:f3}), we used $T=200,000$, $N=100$ and $\eta=40/(t+2000)$.
        For the first example of an imbalanced dataset (Fig.~\ref{fig:f4}A-C), we used a network with 784 visible, 10 label and 400 hidden units with $T=100,000$, $N=100$ and $\eta=20/(t+2000)$.
        For the example of pattern completion from an imbalanced dataset (Fig.~\ref{fig:f4}D-E), we used a network with 784 visible, 10 label and 400 hidden units with $T=200,000$, $N=100$ and $\eta=40/(t+2000)$.
              
    \subsection*{3. Indirect sampling likelihood}
    
        To have a quantitative comparison of mixing between different sampling procedures, we used the indirect sampling likelihood (ISL) method \cite{breuleux2010unlearning, desjardins2010tempered}. 
        The method constructs a non-parametric density estimator to evaluate how close each test example is from any of the generated examples.               
        The likelihood of a test sample $\mathbf{y}$ given a series of generated sample $\{ \mathbf{x}_i \}$ is defined as:       
        \begin{align}
             p(\bs{y}) = \frac{1}{N} \sum_{i=1}^{N} \prod_{j=1}^{d} 
              \beta^{1_{\bs{y}_j = \bs{x}_{ij}}} {(1-\beta)}^{1_{\bs{y}_j \neq \bs{x}_{ij}}}  \quad ,
             \label{eq:isl}
        \end{align}       
        where $N$ is the number of generated samples, $d$ is the dimension of $\bs{y}$ or $\bs{x}_i$ and $\beta$ is a hyperparameter which controls the gain ($\beta$) and punishment ($1-\beta$) to the likelihood when comparing the test sample with the generated sample.
        We set $\beta = 0.95$ to optimize the likelihood; other values ($\beta \in (0.5, 1]$) would rescale the likelihood but without causing qualitative differences.
        
        In Fig.~\ref{fig:f3}B, we plot the mean log-likelihood of 2000 samples from the test set against the number of generated samples.
        The faster increase of the ISL curve for the spiking network is due to better mixing, as the generated samples cover the main modes of the test samples faster (Fig.~\ref{fig:f3}D,E). 
        To provide a frame of reference, we also plotted two additional ISL curves.
        The POM (product of marginals) sampler generated images by sampling each pixel individually from its intensity distribution over the entire training set.
        This sampler preserves the marginal probability distributions for each pixel, but discards any further structure of the image (encoded in correlations between pixel intensities).
        The OPT (optimal) sampler started out with a base set of $10^5$ images generated with AST, from which it randomly picked images sequantially.
        This guarantees optimal mixing for the underlying model, because the base set covers all main modes of the state space, but consecutive samples have no correlation.
     
    \subsection*{4. t-distributed stochastic neighbor embedding}
        
        The tSNE method \cite{maaten2008visualizing} finds a low-dimensional map for a high-dimensional data set, in which the similarity between samples is reflected by their distances in the low-dimensional map.
        Here, we projected the generated digits to a plane to provide an intuitive understanding of the network dynamics and the mixing between different modes (digit classes).
        
        The Euclidean distances between high-dimensional samples $\{\bs{x}_i\}$ are converted into symmetric pairwise similarities   
        \begin{align} 
            p_{ij} = \frac{p_{j|i} + p_{i|j}}{2n} \quad ,
        \end{align}
        where $n$ is the number of samples and $p_{j|i}$ is a conditional probability:
        \begin{align} 
            p_{j|i} = \frac{\exp(- \|\bs{x}_i - \bs{x}_j\|^2 / 2 \sigma_i^2 )} {\sum_{k\neq i} \exp(- \|\bs{x}_i - \bs{x}_k\|^2 / 2 \sigma_i^2)} \quad ,
            \label{eq:pji}
        \end{align}
        with variance $\sigma_i$, which is determined by first defining a so-called perplexity value as the effective number of neighbors of a data point, and then running a binary search.
        For the low-dimensional points $\bs{y}_i$ and $\bs{y}_j$ mapped from the high-dimensional data points $\mathbf{x}_i$ and $\mathbf{x}_j$, the similarity is defined using a t-distribution with one degree of freedom:
        \begin{align}
        q_{ij} = \frac{(1 + \|\bs{y}_i - \bs{y}_j\|^2)^{-1}}
        {\sum_{k \neq l} (1 + \|\bs{y}_k - \bs{y}_l \|^2)^{-1}} \quad .
        \end{align}
        If the mapped points correctly model the similarity between the high-dimensional data points, the similarities $p_{ij}$ and $q_{ij}$ will be equal.
        
        With this motivation, tSNE minimizes the sum of Kullback-Leibler divergences over all data points using a gradient descent method.
        The cost function $C$ is given by
        \begin{align}
        C = D_{KL}(P||Q) = \sum_i \sum_j p_{ij} \log \frac{p_{ij}}{q_{ij}} \quad .
        \end{align}       
        Its gradient with respect to the map point $i$ can then be derived to provide an update of the mapping:
        \begin{align}
        \Delta \bs y_i \propto \frac{\partial C}{\partial \bs{y}_i} = 4 \sum_j (p_{ij}- q_{ij}) (\bs{y}_i - \bs{y}_j) ( 1+ \|\bs{y}_i -\bs{y}_j\|^2)^{-1} \quad .
        \end{align}

\end{document}